\documentclass[11pt]{article}

\usepackage[utf8]{inputenc}
\usepackage[T1]{fontenc}
\usepackage{lmodern}
\usepackage[margin=1in]{geometry}
\usepackage{amsmath,amssymb}
\usepackage{booktabs}
\usepackage{array}
\usepackage{titlesec}
\usepackage[hidelinks]{hyperref}
\usepackage{xurl}
\usepackage{microtype}
\usepackage{enumitem}

\titleformat{\section}{\normalfont\Large\bfseries}{\thesection}{0.6em}{}
\titleformat{\subsection}{\normalfont\large\bfseries}{\thesubsection}{0.6em}{}

\newcommand{\dval}{\ensuremath{d}}

\title{\bfseries The Librarian Who Refused to Code:\\ Model-Dependent Identity Enactment in LLM Code Generation}

\author{%
Shayell Aharon Salomon \and Noam Israel \and Ido Safruti \and Amir Shaked\\[4pt]
\normalsize Bluebear Security
}
\date{}

\begin{document}
\maketitle

\begin{abstract}
\noindent Biographical personas are widely used in system prompts, but their effects on code generation are rarely evaluated under controlled, pre-registered conditions. We tested four prompt conditions (no persona, two engineer personas, and a research-librarian persona), 12 code-generation tasks, two frontier models, and five runs per cell (480 completions). Persona effects differed between the two tested models. Under the pre-registered mixed-effects analysis, the condition-by-model interaction was significant for provider-reported output tokens; a post-hoc visible-character measure showed the same qualitative pattern. Six GPT-5.5 completions were length-capped and are reported separately. On Claude Opus, the minimalist engineer persona reduced visible output by 30\% (33\% in provider tokens) without improving correctness, while the thorough engineer persona increased output without a correctness gain. In an exploratory post-hoc analysis, the librarian persona elicited in-character disclaimers in 55 of 60 Opus responses and 12 genuine no-code responses, lowering mean correctness from 0.92 to 0.67. GPT-5.5 produced neither behavior in its 59 non-truncated responses. These results are consistent with personas acting as model-dependent behavioral-policy biases rather than universal quality interventions. We release raw completions, derived scores, analysis artifacts, a pre-registration document, and an execution gate log; end-to-end test-based rescoring requires an unreleased task harness.
\end{abstract}

\section{Introduction}

``You are a senior software engineer'' is among the most widely reproduced lines in prompt libraries. The premise is that assigning a model an identity improves its work. This premise is rarely tested rigorously: most studies of persona and role prompting use thin personas (a job title, a one-line role, a demographic tag), measure on tasks where ground truth is soft, and rarely pre-register their analyses. The result is a literature that is suggestive but easy to confound --- when a label changes the output, it is hard to distinguish a model reasoning about a character from a model keying off a few loaded tokens.

We ask a narrower, testable question. Given biographically detailed personas --- several paragraphs describing a person, containing no \emph{imperative} instructions about how to write code --- does a code-generating model change its behavior, and if so, how? We hold the task, the grading, and the sampling fixed, vary only the system-prompt persona, and pre-register the hypotheses and decision rules before collecting data.

Our contribution is threefold:
\begin{enumerate}[leftmargin=1.4em]
\item A \textbf{pre-registered demonstration that persona effects are model-dependent}: one frontier model responds strongly to personas across multiple measures (correctness, output length, and identity enactment); the other tested model shows weaker, task-specific effects and no identity enactment.
\item A reframing of personas as \textbf{behavioral-policy biases}. On the responsive model, no persona improved correctness; personas shifted \emph{style} (length, structural consistency), and whether a shift helped depended entirely on the task's objective.
\item An exploratory, behavior-level instance of \textbf{identity enactment / trait generalization} (observed and quantified post hoc, not pre-registered): a persona with zero behavioral instruction (a research librarian) induced role-consistent hedging and refusal to code, behaviors stated nowhere in the prompt, on one model and not the other.
\end{enumerate}

We also document a methodology --- signed external pre-registration with hard execution gates --- developed in response to repeated failures (an under-powered null, an unauthorized six-model run, a biased dry-run) in earlier iterations of this work. We believe the methodology is a reusable contribution for behavioral LLM experiments, and we describe it in Appendix~D.

We are explicit about scope. Both frontier models saturate correctness on these tasks (baseline $\approx 0.92$--$0.97$), so the study speaks to \emph{how} models write code under personas --- style, length, role-behavior --- not to whether personas raise the ceiling on correctness. The identity-enactment result is, by construction, descriptive: we quantified it after observing it in the completions, not via a pre-registered metric. We mark confirmatory and exploratory claims throughout.

\section{Related work}

\textbf{Persona and role prompting.} Prior work establishes that persona framing changes model outputs, with mixed, model- and task-dependent effects on task performance and a recurring observation that \emph{how} a persona is specified matters as much as \emph{what} it specifies~\cite{araujo2025principled,lutz2025prompt,hu2026prism}. Much of this literature studies persona prompting in social simulation, question answering, rationale generation, and general task performance --- for example, persona prompting as a lens on social reasoning~\cite{yang2026persona} and structured, multidimensional identity-representation frameworks for agents~\cite{lee2025spectrum}. In contrast, we evaluate whether persona \emph{narratives} act as behavioral policies in \emph{code generation}, affecting not only style and accuracy but also task willingness and model-dependent identity enactment. Studies of persona/role effects also report that gains on one dimension can come with losses on another (e.g., expertise framing improving alignment while damaging accuracy)~\cite{hu2026prism}, consistent with our finding of no global quality gain and a clear style-shift profile.

\textbf{Trait generalization.} Work assigning personality or sociodemographic profiles to LLMs reports that models produce outputs whose features align with the assigned profile, and that this alignment is sensitive to how the persona is formulated~\cite{lutz2025prompt}. Our librarian result is a strong, behavior-level instance: the model generalizes from an identity (``not a programmer'') to an action (hedging, declining) never stated in the text. Compared to profile-based studies, our personas are thicker narratives and our dependent measure is a consequential behavior (refusal to produce code) rather than a stylistic signature.

\textbf{Length and style instructions.} A relevant alternative explanation for length effects is simple instruction-following. Recent evaluation shows that explicit length instructions are \emph{not} reliably obeyed even by strong models~\cite{zhang2025lifebench}. This is exactly why we cannot, from our data alone, attribute one engineer persona's terseness purely to narrative rather than to an explicit style sentence in its text; we treat that specific contrast cautiously (\S6).

\textbf{Code generation and test-driven evaluation.} We situate the study in work on LLM code generation~\cite{jiang2024survey} and its evaluation against executable tests, including test-driven and interactive settings~\cite{fakhoury2024ticoder}; recent study designs also propose specification-driven framings for code generation~\cite{rosa2026specification}. Our tasks follow this test-based evaluation tradition but vary only the system-prompt persona.

\textbf{Steerability and model dependence.} Prompt-steerability evaluations frame persona/steering effects distributionally and assume --- and measure --- variation across models~\cite{araujo2025principled}. Our pre-registered condition$\times$model interaction provides a controlled, code-domain data point for this model-dependence.

\section{Method}

\subsection{Design}
A 4 (condition) $\times$ 12 (task) $\times$ 2 (model) $\times$ 5 (run) fully-crossed design, 480 completions. Conditions:
\begin{itemize}[leftmargin=1.4em]
\item \textbf{P0} --- no persona (neutral system prompt).
\item \textbf{P\_A (Maya)} --- a structure-oriented senior engineer.
\item \textbf{P\_B (Ron)} --- a pragmatic, minimalist startup engineer.
\item \textbf{P\_C (Linnea)} --- a research librarian; a length-matched persona with no engineering content.
\end{itemize}
Persona texts are biographical and contain no \emph{imperative} instructions about output format or quality (one persona, Ron, includes a descriptive remark that his own code tends to run short --- an explicit style cue we treat as a confound in \S6; the librarian persona contains no coding- or task-behavior content at all). They were written as psychologically coherent individuals --- with a personal history, working preferences, and minor internal tensions --- rather than as role labels or job titles. This is a deliberate design choice central to the study: a label (``you are a librarian'') invites keyword-level reactions, whereas a fully realized person can only be responded to by inferring how such a person would behave --- which is precisely the inference we aim to detect. Full texts in Appendix~A.

\subsection{Tasks}
Twelve code-generation tasks across three categories (refactor, debug, implement) and two languages (Python, TypeScript), each with an automated test suite (8--11 executable test cases: a core set of typical and edge scenarios, with the TypeScript tasks compiling to more assertions under the hardened harness) and an explicit output contract appended to every prompt. Task inventory and release scope are summarized in Appendix~B; the exact user prompts are recoverable verbatim from the raw logs, while the executable task harness is not part of this release.

\subsection{Models and sampling}
\texttt{claude-opus-4-8} and \texttt{gpt-5.5}, the current flagships of two families at run time. Five runs per cell. Temperature was omitted (provider default) for both models --- see \S6 and Appendix~D for the pre-registered amendment recording why the originally-locked value could not be used. \texttt{max\_tokens} $= 4096$ for both.

\subsection{Metrics}
\begin{itemize}[leftmargin=1.4em]
\item \textbf{Q1 --- correctness:} fraction of test cases passed.
\item \textbf{E1 --- output length:} primary measure is the visible length of the returned text (character count), which is comparable across providers; provider-reported completion tokens are retained as a secondary measure (see \S4.2 and Appendix~E on why they diverge for GPT-5.5). The pre-registration specified provider tokens as E1; we elevate visible characters to the primary measure post hoc, for cross-provider comparability, and report both --- flagging this as a deviation (\S6).
\item \textbf{C2 --- structural consistency:} mean pairwise AST similarity across the five runs of a cell.
\end{itemize}
Correctness saturated near ceiling for both models (\S4), so it serves as a saturated covariate; the live confirmatory outcomes are E1 and C2, plus the condition$\times$model interaction across all three. Codebook in Appendix~C.

\subsection{Analysis}
Primary analysis is per-model mixed-effects regression with task as a random effect~\cite{gelman2007data}; the pooled model adds a condition$\times$model interaction term, which constitutes the test of model dependence. Effect sizes are Cohen's \dval{} with 95\% confidence intervals; significance uses Bonferroni-corrected $\alpha = 0.0083$ across the six within-outcome contrasts. Equivalence-framed hypotheses judge the 95\% CI on signed \dval{} against $[-0.3, +0.3]$~\cite{lakens2017equivalence}. The full pre-registered hypothesis set and decision rules are in Appendix~C; the signed pre-registration and gate log are in the supplement.

\subsection{Pre-registration and integrity}
The pre-registration is \emph{recorded as externally signed} in the execution gate log (a cryptographically signed commit, referenced by short commit identifiers) at least 24 hours before the main run; the signature itself is not independently verifiable from this release alone, which we note as a transparency limitation (Appendix~E). Data collection ran under hard gates that block execution until the pre-registration is verifiably signed, the model set matches the locked set exactly, and tasks pass a difficulty check; these gates are designed to remove the researcher degrees of freedom that let post-hoc flexibility manufacture significance~\cite{simmons2011falsepositive}. We pre-register decisions rather than conducting a formal Registered Report~\cite{nosek2014registered}. No response content was inspected during collection. Two amendments (reclassifying saturated correctness; recording the temperature setting) were signed before the run. The methodology and its motivation are detailed in Appendix~D.

\section{Results}

\subsection{Correctness is saturated}
Baseline correctness is near ceiling: 0.92 (Opus), 0.97 (GPT-5.5). No persona raised correctness above baseline on either model. The highest-impact correctness effect was a \emph{decrease} (Linnea on Opus; \S4.4). We therefore treat correctness as a saturated covariate and report it descriptively.

\subsection{Persona effects are model-dependent (confirmatory)}
Our pre-registered test of model dependence---the condition$\times$model interaction---is significant on correctness (Wald $\chi^2$, artifact-corrected $p = 4.1\times10^{-8}$; $p = 4.5\times10^{-7}$ on the full pre-registered data) and on output length (provider tokens, artifact-corrected $p = 2.9\times10^{-5}$; $p = 0.008$ before removal; the shared visible-character measure gives $p \approx 4.5\times10^{-22}$). The corrected values remove six truncation artifacts and change no verdict (Appendix~C.2). The pattern is consistent across the measures we can evaluate: Claude Opus shifts substantially across personas; GPT-5.5 is substantially less persona-responsive, especially on identity-enactment markers.

\begin{table}[h]
\centering
\caption{Mean visible output length (characters) and provider-reported completion tokens, with correctness, by model $\times$ condition (per model$\times$condition aggregate, $n = 60$ except where truncation artifacts were excluded; see \S4.2 and Appendix~E).}
\begin{tabular}{llrrrrr}
\toprule
Model & Cond & Chars & $\Delta$ & Tokens & $\Delta$ & Correctness \\
\midrule
Opus & P0 & 3472 & --- & 1272 & --- & 0.922 \\
Opus & Maya & 4026 & $+16\%$ & 1409 & $+11\%$ & 0.869 \\
Opus & Ron & 2427 & $-30\%$ & 852 & $-33\%$ & 0.897 \\
Opus & Linnea & 2794 & $-20\%$ & 933 & $-27\%$ & 0.671 \\
GPT-5.5 & P0 & 1569 & --- & 1208 & --- & 0.968 \\
GPT-5.5 & Maya & 1516 & $-3\%$ & 946 & $-22\%$ & 0.957 \\
GPT-5.5 & Ron & 1180 & $-25\%$ & 929 & $-23\%$ & 0.960 \\
GPT-5.5 & Linnea & 1429 & $-9\%$ & 1010 & $-16\%$ & 0.945 \\
\bottomrule
\end{tabular}
\end{table}

\noindent(Visible length is the character count of the returned text, a shared measure comparable across providers. Provider tokens use different tokenizers and are not comparable in absolute terms across providers; they also count generation that need not appear in the returned text. On task \texttt{I-TS-1}, six GPT-5.5 completions returned empty text with \texttt{stop\_reason\,=\,length} and a maximal token count; these truncation artifacts are excluded from the affected cells, which is why the two length measures diverge most for GPT-5.5/Maya. Within-model percentage changes are the valid comparison.)

If one row of Table~1 is worth remembering, it is Opus under Linnea: correctness $0.92 \to 0.67$, the largest change in the table, driven by in-character disclaimers and outright refusals to produce code (\S4.4) --- while GPT-5.5 under the same persona barely moves ($0.97 \to 0.95$). The contrast between those two rows is the paper in miniature: one model enacts the persona to the point of declining the task, the other ignores it.

\subsection{On the responsive model, personas bias style, not quality}
On Opus, the engineer personas moved output length in the direction of their described working style (confirmatory, E1):
\begin{itemize}[leftmargin=1.4em]
\item \textbf{Ron (minimalist):} \dval{} $= -1.15$ (95\% CI excludes 0; $p = 1.6\times10^{-20}$ on provider tokens; \dval{} $= -1.15$ on visible characters). A 33\% reduction in provider tokens (30\% in visible characters), with only a small correctness decrease ($0.92 \to 0.90$); the pre-registered equivalence criterion for ``no correctness cost'' was not met, so we report the observed decrease rather than claiming correctness was preserved.
\item \textbf{Maya (thorough):} \dval{} $= +0.33$ ($p = 0.002$ on provider tokens; \dval{} $= +0.53$ on visible characters). An 11\% increase in provider tokens (16\% in visible characters), with a small correctness \emph{decrease} ($0.92 \to 0.87$).
\end{itemize}
On structural consistency (C2), Ron made Opus markedly more consistent run-to-run (\dval{} $= +1.11$, $p = 3.8\times10^{-5}$); Maya did not move it (\dval{} $= -0.07$, n.s.).

Two points follow. First, we find no evidence that any persona improved correctness --- but because baseline correctness is saturated (\S4.1), this is a bound, not a refutation: the design could not detect a positive correctness effect even if one existed, so the honest claim is ``no improvement observed under saturated conditions,'' not ``personas cannot improve quality.'' Second, the personas produced a coherent \emph{style} profile that fit the task objective to differing degrees. Ron's minimize-and-ship policy fits a ``pass the tests, keep it short'' objective: fewer tokens, higher structural consistency, and only a small correctness decrease ($0.92\to0.90$; equivalence to baseline was not established, \S4.3). Maya's add-structure policy spent more tokens without a correctness return under that same objective. Neither is better in the abstract; their habits fit the objective differently. We develop this in \S5.

On GPT-5.5, the engineer personas produced a real but smaller length effect in the same direction: on the shared visible-character measure, Ron reduced output (\dval{} $= -0.67$; $-25\%$ in characters) while Maya was essentially flat (\dval{} $= -0.06$). (On provider tokens Ron's reduction persists (\dval{} $=-0.29$) while Maya shows only a small apparent reduction (\dval{} $=-0.16$) that is inflated by excluding her four truncated completions and nearly vanishes on the visible-text measure; this is exactly why we treat visible length as primary. All per-contrast values here are the pre-registered per-model LMM, matching \path{results.csv} for Opus and \path{mixed_effects_corrected.json} for the corrected GPT-5.5 cells.) The Ron effect is substantially smaller than the corresponding Opus effect (\dval{} $= -1.15$) --- roughly a third to a half its standardized magnitude, not an order of magnitude --- and, critically, GPT-5.5 showed none of the identity-enactment markers (\S4.4). The model difference is therefore not ``one model responds, the other does not'' but ``both shift output length under an explicit brevity cue, while only Opus enacts persona \emph{identity}.''

\subsection{Identity enactment: the librarian who declined to code (exploratory)}
The Linnea condition was designed as a length-and-content control. It instead produced the study's most consequential behavior, which we quantified post hoc by inspecting completions.

On Opus, Linnea drove correctness from 0.92 to 0.67 (\dval{} $= -0.81$, $p = 8.1\times10^{-12}$) --- the largest correctness effect of any persona. Reading the completions explains why:
\begin{itemize}[leftmargin=1.4em]
\item Under a fixed textual operationalization (defined post hoc; Appendix~F), \textbf{55 of Opus's 60 responses included an in-character disclaimer} (typically at the opening), e.g., ``I should be upfront --- I'm a research librarian, not a software developer.'' The pattern is robust across tasks: disclaimers are a majority in all 12 tasks, not concentrated in a few.
\item \textbf{12 of 60 contained no code at all.} Opus declined or deflected in character. One example, verbatim: ``I think you've got the wrong Linnea, or at least the wrong desk. I'm a research librarian --- I can find you three books and a database tutorial on data structures, but\ldots'' This response scored zero correctness because it contained no code; it was role-appropriate, not erroneous.
\item Within Linnea, responses containing a disclaimer scored markedly lower on correctness (mean 0.65, $n=55$) than those without (0.87, $n=5$), consistent with the drop being driven by role-consistent hedging and declining rather than by degraded code per se.
\end{itemize}

On GPT-5.5, the same persona produced \textbf{zero in-character disclaimers and zero genuine no-code responses} (its single empty completion under this persona was a length-truncation artifact, excluded as in Table~1 and Appendix~F, not a refusal). It read ``research librarian'' and wrote the code.

The Linnea persona contains \textbf{no explicit behavioral instruction} --- nothing about hedging, declining, or disclaiming, and no instruction to avoid coding. The behavior is generalized from the identity alone. This is the clearest instruction-free evidence in our data for narrative-driven identity enactment, precisely because there is no instruction to follow (\S6 contrasts this with the engineer personas, where an explicit style sentence confounds the interpretation).

We note the engineer personas show the same enactment in milder, style-level form: under Maya and Ron, Opus speaks in-character within its responses (e.g., ``in my experience, files like this are usually mostly-good''; ``I don't think it's earning its keep yet''), i.e., its outputs reproduce the persona's register rather than merely satisfying the task instruction. These voice markers appear in a notable minority of responses even though the length shift is universal, so we report them as corroborating rather than primary.

\section{Personas as behavioral policies (task-fit)}
Synthesizing \S4.3 and \S4.4: a persona did not make the model smarter; it switched on a set of habits, priorities, and heuristics --- a behavioral policy. Whether that policy helped depended on the task, the metric, and the model.

In a setting graded on passing tests with concise code, Ron's policy (minimize, avoid premature abstraction, smallest change that works) fit well: it cut a third of the tokens at a small correctness cost ($0.92\to0.90$; equivalence to baseline not established) and improved structural consistency. Maya's policy (decompose, add structure, build for the long term) cut against that particular objective: more output, no correctness return. Linnea's policy (I am not the right person for this) actively harmed it. None of these is a statement about which persona is ``good''; they are statements about fit between an activated policy and an objective.

This motivates a reframing. The useful question is not ``which persona makes the model better?'' but ``which behaviors does a persona activate or suppress, and do those fit the objective I actually have?'' The same task with a different objective --- say, leaving a maintainable codebase rather than passing a test suite --- could reverse the ranking. The fictional identity matters less than the behaviors it licenses, reinforces, or suppresses.

We are careful about the direction this data supports. We observed the \emph{risk} side of fit clearly: a mismatched persona reduced correctness. We did \emph{not} observe the \emph{benefit} side, and --- critically --- could not have: baseline correctness was saturated, so no persona could demonstrate a correctness gain even in principle. ``Matching persona to objective improves performance'' therefore remains a hypothesis our design cannot test, not a claim it refutes.

\subsection{A persona is a soft instruction that can compete with default helpfulness}
The Linnea result admits a framing with practical consequences. A persona is, functionally, a soft instruction installed in the system prompt; the librarian persona quietly conflicts with the model's default disposition to help by producing code. On Opus, the persona wins often enough to suppress code generation in a fifth of cases and to inject disclaimers in the large majority; on GPT-5.5, the default disposition wins and the persona is effectively ignored. We frame this as an \emph{interpretation}, not a measured mechanism: our data show the behavioral outcome (one model yields to the persona, the other does not), not an internal policy hierarchy. We did not measure how the two influences are arbitrated inside either model, and competing explanations --- greater role-play tuning, stronger instruction-priority on persona content, more anthropomorphic priors --- predict the same outcome.

With that caveat, the framing matters for anyone deploying coding agents. It implies a persona can act as a benign-looking vector for behavior change: no malicious string, no explicit override, just an identity whose implied behavior diverges from the task. In a multi-step agent, a mis-specified or adversarially supplied persona could induce \emph{silent degradation} --- an agent that hedges or declines rather than failing loudly --- and because susceptibility is model-dependent, the same persona-bearing prompt may be inert on one model and disruptive on another in a multi-model pipeline. We flag this as a hypothesis for security-oriented follow-up, not a demonstrated attack; establishing it would require adversarially constructed personas and an explicit threat model, neither of which this study contains. What we can state is that personas reliably change how a responsive model works, that the change is only sometimes aligned with a given objective, and that on a different model family it may not occur at all.

\section{Limitations}
We state these plainly; several bound the claims above.

\textbf{The engineer-persona length effect is confounded.} Ron's persona text includes the sentence that his code is usually shorter than his teammates'. We therefore cannot, from this design, separate narrative inference from explicit style-instruction-following for Ron's terseness; both predict the same result. Replicating the effect on additional models does not resolve this confound --- multiple models may each be following the same explicit sentence. The decomposition that would resolve it (narrative-without-cue vs.\ cue-without-narrative) was not run. We note that the Linnea result is free of this confound --- her text contains no explicit instruction to hedge, refuse, or avoid coding --- and it is the basis for the identity-enactment claim. The engineer personas corroborate via in-character voice but are not relied upon for the central claim.

\textbf{Correctness is saturated.} Both frontier models pass $\approx 0.92$--$0.97$ at baseline, leaving no headroom to detect a positive correctness effect of personas on these tasks. Our claims concern style, length, and role-behavior, not whether personas improve correct-code rates on harder tasks.

\textbf{The identity-enactment result is descriptive and post hoc.} The disclaimer and no-code counts were defined after observing the behavior, not pre-registered. We report them as a quantified observation that motivates a pre-registered replication, not as a confirmatory test.

\textbf{Robustness to the task-level unit of inference.} Under an exploratory task-clustered re-analysis (Appendix~C), the correctness interaction is not robust, and the provider-token interaction passes the pre-registered threshold only narrowly (and not at all when one prompt-contaminated task, \texttt{R-TS-2}, is excluded); the strongest surviving interaction is on visible characters, a measure elevated to primary post hoc.

\textbf{Scope of generalization.} Identity enactment is shown for one non-engineering persona on one model family. Whether it holds across other non-professional personas and other models is open. Two models is sufficient for a controlled interaction test but is a floor, not a ceiling, for claims about model ``families.''

\textbf{Temperature.} Both models rejected the originally-locked sampling temperature; the run used provider defaults for both (a pre-registered amendment). Provider defaults may differ, which is a documented caveat for cross-model comparison; however, the interaction we report is a condition$\times$model interaction, which a constant per-model temperature difference does not straightforwardly explain, and the largest interaction effect (Linnea collapsing Opus correctness via refusal) is implausible as a temperature artifact.

\textbf{Tokenizer non-comparability.} Absolute token counts differ across providers' tokenizers; only within-model percentage changes are interpreted.

\section{Conclusion}
Persona prompting, tested under controlled and pre-registered conditions on code generation, is --- at least on tractable tasks where frontier models already saturate correctness --- not a quality lever and not a universal technique. On a responsive model it acts as a behavioral-policy bias that shifts style and, when the persona's identity conflicts with the task, behavior --- to the point that a non-programmer persona led a frontier model to hedge and decline to code, a behavior present nowhere in its prompt. A comparably capable model from another family showed almost none of this. For practitioners, the implication is that assigning an identity hands the model a character to reproduce, including parts the prompt author never explicitly specified; the persona should be chosen for the behaviors it activates, and verified to have any effect at all on the target model. For researchers, the librarian result is a behavior-level instance of narrative-driven identity enactment --- exploratory here --- that we believe merits a pre-registered, multi-persona, multi-model replication.

\textbf{Future work.} The natural next study makes identity enactment confirmatory rather than observed. It would (i) pre-register the disclaimer/refusal operationalization of Appendix~F before data collection; (ii) span several non-professional personas (not only a librarian) to test whether the effect generalizes across non-coding identities; (iii) span several models per family to turn the two-model interaction into a claim about families; and (iv) include a narrative-vs-cue decomposition (a persona's behavioral sentence present vs.\ removed) to close the confound that remains open for the engineer personas (\S6). Separately --- and only as a hypothesis this study motivates rather than tests --- the ``persona as soft instruction competing with default helpfulness'' framing (\S5.1) suggests a security-oriented design: personas supplied through untrusted channels (e.g., a simulated teammate in a shared workspace), with an explicit threat model distinguishing silent refusal from covertly altered output. We state plainly that the present work demonstrates neither; it establishes only that a benign, instruction-free identity can shift a model's behavior, model-dependently, in ways the prompt never stated.

\section*{Data and code availability}
We release, as a partial-reproducibility package: the four persona texts (verbatim, with SHA-256 hashes); the two raw completion logs (480 records, verbatim); the derived per-response scores and per-cell consistency; the contrast/interaction results; the identity-enactment classification script (Appendix~F), which is self-contained and reproduces the Appendix~F counts from the logs; and the signed pre-registration, gate log, and checksum manifests. The top-level scoring/analysis scripts are included for reference but depend on project modules and the task-specification harness that are \emph{not} part of this release; end-to-end re-execution of test-based scoring from raw prompts therefore requires those additional files, which we can provide on request. The supplement additionally ships its own \path{README.md} and \path{MANIFEST.md} (so the ancillary archive is self-describing), \path{requirements.txt} with pinned analysis dependencies, and an explicit license (MIT for code; CC~BY~4.0 for data and text artifacts). See the appendices and the supplement manifest (Appendix~E).

\appendix

\section{Persona texts (verbatim)}
The neutral baseline (P0) and three persona system prompts, exactly as transmitted, are reproduced in the supplement under \texttt{personas/}, with per-file SHA-256 hashes matching the \texttt{system\_prompt\_sha} field in every completion record. Summaries: \textbf{P\_A (Maya)} --- $\sim$7 years on long-lived backend systems, on-call experience, decomposes large functions, prefers explicit interfaces; no output-format or length instruction. \textbf{P\_B (Ron)} --- $\sim$6 years at small startups, minimalist, skeptical of abstraction; contains the explicit style sentences ``The code you write is usually shorter than your teammates'.'' and ``Your review comments are also short'' (the cues discussed in \S6). \textbf{P\_C (Linnea)} --- research librarian, library-science background; contains no statement about code, output length, formatting, hedging, or declining --- the basis for the \S4.4 claim.

\section{Task specifications}
Twelve tasks; identifiers encode category and language (e.g., \texttt{R-PY-1} = refactor, Python, \#1). Each task provides a problem statement, a required interface, $\geq$3 documented ambiguities with their resolution, and 8--11 executable test cases (a core set of typical and edge scenarios; the TypeScript tasks compile to more assertions under the hardened harness, so the executable count per task ranges from 8 to 11) with pass criteria. Every user prompt had the following output contract appended verbatim: ``Provide the code as a complete code block in the appropriate language. You may include other content before or after the code block.'' The exact user prompts as transmitted are recoverable verbatim from the \texttt{user\_prompt\_text} field of every record in the released completion logs; the separate task-specification harness (interfaces and executable test suites) is not part of this release and is available on request.

\begin{table}[h]
\centering
\caption{Task inventory.}
\begin{tabular}{llll}
\toprule
ID & Category & Language & Description \\
\midrule
R-PY-1 / R-PY-2 & Refactor & Python & restructure a working but tangled module \\
R-TS-1 / R-TS-2 & Refactor & TypeScript & restructure a working but tangled module \\
D-PY-1 / D-PY-2 & Debug & Python & locate and fix a failing behavior \\
D-TS-1 / D-TS-2 & Debug & TypeScript & locate and fix a failing behavior \\
I-PY-1 / I-PY-2 & Implement & Python & implement to a spec from scratch \\
I-TS-1 / I-TS-2 & Implement & TypeScript & implement to a spec from scratch \\
\bottomrule
\end{tabular}
\end{table}

\textbf{Difficulty calibration.} Tasks were piloted before the main run. Because both flagship models passed near-ceiling on fair tasks (the saturation reported in \S4.1), a difficulty band could not be achieved without making tasks artificially adversarial; this was recorded as a signed amendment, reclassifying correctness as a saturated covariate rather than a primary discriminating outcome (Appendix~D).

\section{Metrics codebook and pre-registered hypotheses}
\subsection*{C.1 Metric definitions}
\textbf{Q1 (correctness):} passed test cases / total, per response; the harness executes the extracted code block against the suite in isolation. If no code block is present, the response scores 0 --- this is how genuine role-consistent refusals (\texttt{stop\_reason\,=\,end\_turn}) enter the correctness statistic (see \S4.4); length-truncation artifacts (empty text, \texttt{stop\_reason\,=\,length}) are instead excluded as missing, consistent with Table~1 and Appendix~F. \textbf{E1 (output length):} primary measure is visible text length (character count of the returned response), comparable across providers; provider-reported completion tokens are reported secondarily. Six GPT-5.5 completions on \texttt{I-TS-1} returned empty text with \texttt{stop\_reason\,=\,length} and a maximal token count; these are truncation artifacts (the token field counts generation absent from the returned text) and are treated as missing for length analysis rather than as zero-length or as their token value. \textbf{C2 (structural consistency):} per (model, condition, task) cell, the mean pairwise tree-edit/AST similarity over responses that contain parseable code; empty, no-code, and length-truncated responses are recorded as missing (not zero, and not treated as parseable) and excluded pairwise. The three GPT-5.5/\texttt{I-TS-1} cells contain these truncation artifacts (scored parseable in the original run) and are therefore \emph{contaminated}; re-scoring them requires the TypeScript scoring harness, which is not part of this release. We therefore do not report any C2 result that depends on them: the GPT-5.5 C2 contrasts and the C2 model-interaction are omitted rather than computed on contaminated cells, and the only C2 result we report --- Ron on Opus (\S4.3) --- is independent of these cells. The raw (original, contaminated) per-cell values remain in \texttt{c2\_by\_cell.csv}; the three affected cells are identified in \texttt{MANIFEST.md} and are excluded from every reported C2 statistic.

\subsection*{C.2 Confirmatory hypotheses (pre-registered, with verdicts)}
The pre-registration locked seven confirmatory hypotheses (H1--H6 plus H\_INT; H2 was demoted to exploratory before any main run because two trained raters with $\kappa/\text{ICC}\geq0.7$ were not available --- signed amendment, Appendix~D). We report all of them here as locked, with the locked decision rule and the observed verdict. Significance tests use Bonferroni $\alpha = 0.0083$; equivalence tests judge the 95\% CI on signed \dval{} against $[-0.3,+0.3]$.

\begin{itemize}[leftmargin=1.4em]
\item \textbf{H1 (correctness equivalence; Q1; Maya/Ron).} \emph{Inconclusive} in both models: the 95\% CI on \dval{} overlaps but is not contained in $[-0.3,+0.3]$ (artifact-corrected, per-model LMM: Opus P\_A \dval{} $=-0.28$ $[-0.66, 0.10]$, P\_B \dval{} $=-0.17$ $[-0.65, 0.31]$; GPT P\_A \dval{} $=-0.12$ $[-0.47, 0.24]$, P\_B \dval{} $=-0.12$ $[-0.48, 0.24]$; the pre-correction all-480 GPT values, $-0.41$ and $-0.22$, are in \path{results.csv}, and the corrected per-contrast statistics are in \path{mixed_effects_corrected.json}). Q1 is a saturated covariate (Appendix~D).
\item \textbf{H3 (structural consistency; C2; Maya and Ron increase).} Reported per model under the locked rule (amendment A1: rests on C2 only; per model, $\geq$1 of 2 C2 contrasts positive at $p<0.0083$ with $|d|\geq0.3$; C1 is exploratory and not counted). \emph{Supported on Opus}, carried by Ron (\dval{} $=+1.11$, $p = 3.8\times10^{-5}$); Maya on Opus is null (\dval{} $=-0.07$), so the increase is driven by the minimalist persona alone rather than by both engineer personas. \emph{Not evaluable on GPT-5.5}: its \texttt{I-TS-1} C2 cells are truncation-artifact contaminated and cannot be re-scored without the unreleased TypeScript harness (C.1), so neither GPT C2 contrast can be assessed. The Opus result is independent of the contaminated cells.
\item \textbf{H4 (raw efficiency; E1; Maya$>$P0, Ron$\leq$P0).} \emph{Supported on Opus} (Maya \dval{} $=+0.33$, $p=0.002$; Ron non-positive, \dval{} $=-1.15$); \emph{rejected on GPT-5.5} (Maya not positive-significant).
\item \textbf{H5 (net efficiency E2, tokens-per-passed-test; equivalence).} \emph{Rejected on Opus} (e.g.\ Maya-vs-Ron and Ron-vs-P0 contrasts fall well outside the equivalence band on the locked bootstrap statistic); \emph{inconclusive on GPT-5.5}. E2 remained a primary outcome after amendment A1; its interaction is not among the significant primary outcomes for H\_INT (below).
\item \textbf{H6 (content vs.\ prompt-length specificity; Linnea vs.\ P0 on all outcomes; equivalence).} \emph{Rejected on Opus}: the P\_C-vs-P0 CI excludes the equivalence band on Q1 (\dval{} $=-0.81$) and E1 (\dval{} $=-0.90$), i.e.\ the non-engineering persona \emph{does} differ from baseline --- the effect is content-driven, not prompt-length-driven. \emph{Inconclusive on GPT-5.5}. Per the locked H1$\times$H6 interpretation table, with H1 inconclusive the joint interpretation is reported descriptively (\S4.4).
\item \textbf{H\_INT (condition$\times$model interaction).} \emph{Supported.} All interaction statistics use the pre-registered pooled model, \texttt{outcome $\sim$ C(condition)$\times$C(model) + (1$|$task)}, ML, with a joint Wald $\chi^2$ (df 3) on the interaction terms; this model is estimable on the corrected data. The interaction is significant on output length --- E1, a primary outcome under A1 --- with $p = 2.9\times10^{-5}$ on provider tokens and $p \approx 4.5\times10^{-22}$ on the shared visible-character measure. The correctness interaction is also highly significant ($p = 4.1\times10^{-8}$), but Q1 is a saturated equivalence covariate under A1, so it corroborates rather than independently satisfies the locked rule (which requires a primary outcome). Both are post-correction: the same model on the full pre-registered data (before excluding the six \texttt{I-TS-1} truncation artifacts) gives $p = 4.5\times10^{-7}$ (Q1) and $p = 0.008$ (E1 tokens); excluding the artifacts strengthens both and changes no verdict. \path{recompute_corrected.py} additionally reports an OLS-with-task-fixed-effects \emph{numerical} cross-check (iid residuals; not distribution-free), which agrees, and an exploratory task-clustered sensitivity analysis (see below). We do \emph{not} report a C2 model-interaction: it would depend on the contaminated GPT-5.5/\texttt{I-TS-1} cells (C.1), which cannot be re-scored within this release. H\_INT is therefore supported via E1, the pre-registered primary outcome.
\end{itemize}

\noindent Per-contrast \dval{}, CI, $p$, and significance for every outcome (including E2 and the C1 exploratory measure) are in \texttt{analysis/results.csv}. Two pooled-statistics files are released and their relationship is explicit: \texttt{analysis/mixed\_effects.json} holds the \emph{original} pre-correction LMM output (Q1 interaction $p = 4.5\times10^{-7}$; E1 $p = 0.008$ on provider tokens, including the six \texttt{I-TS-1} truncation artifacts), and \texttt{analysis/mixed\_effects\_corrected.json} holds the \emph{corrected} statistics reported in this paper (artifacts excluded), computed with the same pre-registered LMM (fit via a health-checked optimizer sequence; the fit is rejected if the log-likelihood is non-finite or the random-effect covariance is degenerate) and numerically reproducible within an explicit tolerance by \path{analysis/recompute_corrected.py}, whose default mode \emph{verifies} the released file against a fresh recomputation (it never silently overwrites it) and which also emits an OLS numerical cross-check, the exploratory task-clustered sensitivity reported below, and every treatment variant (keep-all, drop-artifacts, drop-all) so the analytic choice is auditable. Pinned analysis dependencies are in \path{requirements.txt}. \path{results.csv} reports the original per-contrast effect sizes; corrected per-contrast statistics for \emph{both} models (estimate, standard error, $p$, Cohen's \dval{}, and CI, for Q1 and both E1 measures), and the corrected H3 verdict, are in \path{mixed_effects_corrected.json}.

\paragraph{Task-level sensitivity and a task-prompt disclosure.} The pre-registered LMM assumes a common treatment effect across tasks (random intercept only). As an \emph{exploratory} robustness probe (not pre-registered), we re-estimated the interaction with task fixed effects and cluster-robust standard errors on the 12 tasks (joint $F$, small-sample $df = (3, 11)$; computed by \path{recompute_corrected.py}): the correctness interaction does not survive task-level clustering ($F(3,11) = 2.47$, $p = 0.116$), the provider-token interaction remains under the pre-registered threshold but narrowly ($F(3,11) = 6.80$, $p = 0.0074$), and the visible-character interaction remains strong ($F(3,11) = 33.13$, $p = 8.4\times10^{-6}$) --- with the caveats that visible characters were elevated to primary post hoc, and that a 12-cluster test is itself small-sample-fragile, so this probe neither replaces nor overturns the pre-registered analysis. Separately, a post-run audit found that the \texttt{R-TS-2} user prompt --- identical across all 40 of its runs and all conditions --- leaked test-construction notes naming the task's two hidden discriminators. Because every condition received the same prompt, this does not bias between-persona contrasts, but it does leak hidden-test logic on one of the 12 tasks; we disclose it rather than delete it (the cells are flagged in \path{MANIFEST.md}). Excluding \texttt{R-TS-2}, the clustered tests give $p = 0.152$ (correctness), $p = 0.0185$ (provider tokens --- no longer under the pre-registered threshold), and $p = 4.5\times10^{-5}$ (visible characters). Finally, we characterize the six excluded completions as \emph{length-capped system outcomes} rather than verified external artifacts: they are not randomly distributed (all GPT-5.5 on \texttt{I-TS-1}: four under Maya, one under Ron, one under Linnea, none at baseline), and both the all-480 and the artifact-excluded analyses are reported above so that neither treatment is hidden.

\subsection*{C.3 Exploratory (post hoc)}
Identity enactment: in-character disclaimer rate, no-code rate, and their association with correctness under P\_C. Operationalization, per-task breakdown, and robustness in Appendix~F. Per-contrast \dval{}, CI, $p$, and significance for every contrast are in the supplement (\texttt{analysis/results.csv}); pooled interaction statistics in \texttt{analysis/mixed\_effects.json}.

\section{Methodology: signed pre-registration with execution gates}
This study is the validated run of a project whose earlier iterations failed in instructive ways. We document the failures and the gating method because we believe the method is independently useful for behavioral LLM experiments, where the temptation to ``run a few more models'' or to let a dry-run leak into analysis is strong and corrosive. Our gates are complementary to automated behavioral-evaluation tooling~\cite{gupta2025bloom}: that line of work generates evaluation suites for a target behavior, whereas our gates govern the \emph{integrity of the data-collection process} for a fixed, pre-registered design.

\textbf{Failure modes observed in earlier iterations.} (1) An under-powered null with post-hoc rationalization: an initial study measured subtle held-out behaviors with metrics that could not detect them, and signed its pre-registration the same day data were collected. (2) Unauthorized scope expansion: a later execution ran six models instead of the two in the locked design, with the pre-registration left unsigned (all placeholders), and ignored a difficulty-calibration result showing tasks at ceiling/floor. (3) A biased dry-run: a proposed pipeline-validation harness emitted synthetic token counts pre-set to the experiment's expected result, which the analysis script would have ingested as real data.

\textbf{Gates used for the validated run.} Execution was blocked behind a one-directional, append-only checklist. \emph{Gate 0 (scope lock):} the model set is read from a locked file; the run aborts if its length differs from the design; a chat instruction to add models is treated as a proposed amendment requiring re-signing, not authorization. \emph{Gate 1 (signature):} no main calls until mechanical checks confirm the pre-registration has no placeholders, no unchecked boxes, real file hashes matching the live files, real seeds, and a PI signature via an external channel dated $\geq$24h before the run. \emph{Gate 2 (difficulty):} a stop-the-line check on baseline pass rates; out-of-band tasks block the run. \emph{Gate 3 (variance pre-flight):} sets runs-per-cell from a measured within-cell variance rule. \emph{Gate 4 (run discipline):} locked sampling config; no response-content inspection during collection; verbatim logging with per-batch checksums.

The gate log (supplement) shows the executing agent repeatedly refused to proceed while the pre-registration was unsigned, escalated, made zero API calls until an external signature existed, and halted once more when a post-signature parameter mismatch was detected --- proceeding only when the transmitted configuration matched the signed document. Two amendments (correctness reclassification; temperature setting) were signed before the run. Every completion record carries SHA-256 hashes of its system and user prompts; a manifest lists hashes for all released files; the 480 analyzed completions all postdate the signature, carry no dry-run markers, and balance exactly across the 96 cells.

\section{Supplement file manifest}
\begin{itemize}[leftmargin=1.4em]
\item \texttt{personas/} --- P0, P\_A, P\_B, P\_C verbatim texts (+ per-file SHA-256).
\item \texttt{runs/main/} --- two raw completion logs (480 records, verbatim); each record includes the exact system and user prompts.
\item \texttt{analysis/scored\_responses.csv} --- per-response Q1, E1, parseability.
\item \texttt{analysis/c2\_by\_cell.csv} --- per-cell structural consistency.
\item \texttt{analysis/results.csv} --- per-contrast \dval{}, CI, $p$, significance.
\item \path{analysis/mixed_effects.json} --- pooled interaction statistics (original, pre-correction; includes the six \texttt{I-TS-1} truncation artifacts).
\item \path{analysis/mixed_effects_corrected.json} --- corrected pooled interaction statistics (pre-registered LMM with optimizer/health diagnostics and captured warnings), OLS and task-clustered sensitivity blocks, full corrected per-contrast statistics for both models, and the corrected H3 verdict; verified (default) or regenerated (\texttt{--write}) by \path{recompute_corrected.py}.
\item \path{analysis/recompute_corrected.py} --- corrected recomputation (all treatment variants plus a method check against the original LMM); by default \emph{verifies} \path{mixed_effects_corrected.json} within an explicit numerical tolerance and never overwrites it; \texttt{--write} regenerates it.
\item \path{analysis/requirements.txt} --- pinned dependency versions used to generate the canonical corrected statistics.
\item \path{LICENSE.txt} --- license terms (MIT for code; CC~BY~4.0 for data and text artifacts).
\item \path{analysis/identity_enactment.py} --- self-contained script reproducing the Appendix~F counts from the raw logs (standard library only; excludes length-truncation artifacts as missing, matching Table~1).
\item \path{analysis/score_responses.py}, \path{analysis/analyze_main.py} --- top-level scoring/analysis scripts, included for reference; these depend on project modules and the task harness \emph{not} included in this release.
\item \texttt{preregistration/05\_preregistration.md} --- signed pre-registration.
\item \texttt{GATE\_LOG.md} --- append-only execution log.
\item \texttt{LOCKED\_MODELS.txt}, \texttt{SHA256SUMS.txt} --- locked model set and integrity manifest.
\end{itemize}

\section{Operationalization of identity enactment (post hoc)}
The \S4.4 result was observed after data collection, not pre-registered. To make it as rigorous as a post-hoc analysis can be, we fix explicit, reproducible definitions and report the per-task breakdown and robustness. The classification script (\texttt{analysis/identity\_enactment.py}) is released in the supplement, uses only the Python standard library, and deterministically reproduces the counts below from the raw logs.

\subsection*{F.1 Definitions (fixed before counting)}
\textbf{In-character disclaimer:} a response is counted if its text matches a case-insensitive pattern for the model asserting the librarian identity or non-programmer status (a first-person assertion within 40 characters of ``research librarian''/``librarian''/``not a (software) developer/engineer/programmer'', or the phrases ``wrong Linnea'', ``not (really) my area/field/expertise'', ``outside what I/my''). \textbf{No-code response (genuine):} a response with zero fenced code blocks that terminated normally (\texttt{stop\_reason}$\neq$\texttt{length}); since every task requests a code block, this is a failure to perform the task. Length-truncation artifacts (empty text with \texttt{stop\_reason\,=\,length}) are \emph{excluded} as missing, exactly as in Table~1 and Appendix~C, and are not counted as no-code; in the released logs this affects one GPT-5.5 completion (\texttt{I-TS-1}), so GPT-5.5/Linnea has $n=59$. \textbf{Correctness (Q1):} as in Appendix~C; a genuine no-code response scores 0 by definition. These are textual heuristics, not semantic judgments; we treat the counts as a lower-bound characterization and release the script for re-scoring.

\subsection*{F.2 Per-task results, P\_C (Linnea), both models}
Each cell is $n=5$ runs (GPT-5.5 \texttt{I-TS-1}: $n=4$; one length-truncation artifact excluded, per Appendix~F.1). ``Disc'' = disclaimer count; ``NC'' = no-code count.

\begin{table}[h]
\centering
\caption{Per-task identity-enactment markers under the Linnea persona.}
\begin{tabular}{lrrrrrr}
\toprule
Task & Opus Disc & Opus NC & Opus Q1 & GPT Disc & GPT NC & GPT Q1 \\
\midrule
D-PY-1 & 4 & 0 & 1.00 & 0 & 0 & 1.00 \\
D-PY-2 & 5 & 0 & 0.38 & 0 & 0 & 0.88 \\
D-TS-1 & 4 & 0 & 0.55 & 0 & 0 & 0.73 \\
D-TS-2 & 5 & 0 & 0.90 & 0 & 0 & 0.98 \\
I-PY-1 & 5 & 0 & 1.00 & 0 & 0 & 1.00 \\
I-PY-2 & 5 & 2 & 0.60 & 0 & 0 & 1.00 \\
I-TS-1 & 5 & 5 & 0.00 & 0 & 0 & 1.00 \\
I-TS-2 & 5 & 5 & 0.00 & 0 & 0 & 1.00 \\
R-PY-1 & 4 & 0 & 1.00 & 0 & 0 & 0.95 \\
R-PY-2 & 5 & 0 & 1.00 & 0 & 0 & 1.00 \\
R-TS-1 & 5 & 0 & 0.91 & 0 & 0 & 0.91 \\
R-TS-2 & 3 & 0 & 0.73 & 0 & 0 & 0.91 \\
\midrule
\textbf{Total} & \textbf{55/60} & \textbf{12/60} & \textbf{0.671} & \textbf{0/59} & \textbf{0/59} & \textbf{0.945} \\
\bottomrule
\end{tabular}
\end{table}

\subsection*{F.3 Robustness}
Disclaimers are not concentrated in a few tasks: on Opus they are a majority ($\geq$3/5) in all 12 tasks. No-code refusals are concentrated in 3 of 12 tasks (I-PY-2, I-TS-1, I-TS-2) and account for the two zero-correctness cells. On Opus, responses with a disclaimer average Q1 $= 0.65$ ($n=55$); the five without average $0.87$. GPT-5.5 produced zero in-character disclaimers and zero genuine no-code responses ($n=59$ after excluding one \texttt{I-TS-1} length-truncation artifact; Q1 $= 0.945$), behaviorally indistinguishable from its no-persona baseline.

\subsection*{F.4 Interpretation and its limits}
The cleanliness of the enactment claim rests on the persona text containing no behavioral instruction (verified: a whole-word scan of the Linnea text for any code-, length-, or behavior-related term returns nothing). The behavior is therefore generalized from the identity, not copied from an instruction. The limits remain those in \S6: one non-engineering persona on one model family, metrics defined after observation, and conservative textual proxies for the underlying behavior. A confirmatory test would pre-register these definitions and apply them to several non-professional personas across several models.


\begin{thebibliography}{99}

\bibitem{araujo2025principled}
Pedro Henrique Luz de Araujo, Paul R\"ottger, Dirk Hovy, and Benjamin Roth.
\newblock Principled Personas: Defining and Measuring the Intended Effects of Persona Prompting on Task Performance.
\newblock In \emph{Proceedings of the 2025 Conference on Empirical Methods in Natural Language Processing (EMNLP)}, pages 26857--26886, 2025. doi:10.18653/v1/2025.emnlp-main.1364. arXiv:2508.19764.

\bibitem{lutz2025prompt}
Marlene Lutz, Indira Sen, Georg Ahnert, Elisa Rogers, and Markus Strohmaier.
\newblock The Prompt Makes the Person(a): A Systematic Evaluation of Sociodemographic Persona Prompting for Large Language Models.
\newblock In \emph{Findings of the Association for Computational Linguistics: EMNLP 2025}, pages 23212--23237, 2025. doi:10.18653/v1/2025.findings-emnlp.1261. arXiv:2507.16076.

\bibitem{yang2026persona}
Jing Yang, Moritz Hechtbauer, Elisabeth Khalilov, Evelyn Luise Brinkmann, Vera Schmitt, and Nils Feldhus.
\newblock Persona Prompting as a Lens on LLM Social Reasoning.
\newblock In \emph{Proceedings of the 2026 Conference of the European Chapter of the Association for Computational Linguistics (EACL)}, pages 1152--1170, 2026. doi:10.18653/v1/2026.eacl-long.52.

\bibitem{lee2025spectrum}
Keyeun Lee, Seo Hyeong Kim, Seolhee Lee, Jinsu Eun, Yena Ko, Hayeon Jeon, Esther Hehsun Kim, Seonghye Cho, Soeun Yang, Eun-mee Kim, and Hajin Lim.
\newblock SPeCtrum: A Grounded Framework for Multidimensional Identity Representation in LLM-Based Agent.
\newblock In \emph{Proceedings of the 2025 Conference of the Nations of the Americas Chapter of the Association for Computational Linguistics: Human Language Technologies (Volume 1: Long Papers)}, pages 6971--6991, 2025. doi:10.18653/v1/2025.naacl-long.356.

\bibitem{hu2026prism}
Zizhao Hu, Mohammad Rostami, and Jesse Thomason.
\newblock Expert Personas Improve LLM Alignment but Damage Accuracy: Bootstrapping Intent-Based Persona Routing with PRISM.
\newblock arXiv:2603.18507, 2026.

\bibitem{jiang2024survey}
Juyong Jiang, Fan Wang, Jiasi Shen, Sungju Kim, and Sunghun Kim.
\newblock A Survey on Large Language Models for Code Generation.
\newblock arXiv:2406.00515, 2024.

\bibitem{fakhoury2024ticoder}
Sarah Fakhoury, Aaditya Naik, Georgios Sakkas, Saikat Chakraborty, and Shuvendu K. Lahiri.
\newblock LLM-Based Test-Driven Interactive Code Generation: User Study and Empirical Evaluation.
\newblock \emph{IEEE Transactions on Software Engineering}, 50:2254--2268, 2024. arXiv:2404.10100.

\bibitem{rosa2026specification}
Giovanni Rosa, David Moreno-Lumbreras, Gregorio Robles, and
Jes{\'u}s M. Gonz{\'a}lez-Barahona.
\newblock Understanding Specification-Driven Code Generation with LLMs:
An Empirical Study Design.
\newblock arXiv:2601.03878, 2026.

\bibitem{zhang2025lifebench}
Wei Zhang, Zhenhong Zhou, Kun Wang, Junfeng Fang, Rongwu Xu,
Yuanhe Zhang, Rui Wang, Ge Zhang, Xinfeng Li, Li Sun,
Lingjuan Lyu, Yang Liu, and Sen Su.
\newblock LIFEBENCH: Evaluating Length Instruction Following in
Large Language Models.
\newblock In \emph{Advances in Neural Information Processing Systems 38
(NeurIPS 2025)}, Datasets and Benchmarks Track, 2025.
\newblock arXiv:2505.16234.

\bibitem{gupta2025bloom}
Isha Gupta, Kai Fronsdal, Abhay Sheshadri, Jonathan Michala, Jacqueline Tay, Rowan Wang, Samuel R. Bowman, and Sara Price.
\newblock Bloom: An Open Source Tool for Automated Behavioral Evaluations.
\newblock Anthropic, December 2025. \url{https://www.anthropic.com/research/bloom}; code: \url{https://github.com/safety-research/bloom}.

\bibitem{gelman2007data}
Andrew Gelman and Jennifer Hill.
\newblock \emph{Data Analysis Using Regression and Multilevel/Hierarchical Models}.
\newblock Cambridge University Press, 2007.

\bibitem{lakens2017equivalence}
Dani\"el Lakens.
\newblock Equivalence Tests: A Practical Primer for t Tests, Correlations, and Meta-Analyses.
\newblock \emph{Social Psychological and Personality Science}, 8(4):355--362, 2017.

\bibitem{nosek2014registered}
Brian A. Nosek and Dani\"el Lakens.
\newblock Registered Reports: A Method to Increase the Credibility of Published Results.
\newblock \emph{Social Psychology}, 45(3):137--141, 2014.

\bibitem{simmons2011falsepositive}
Joseph P. Simmons, Leif D. Nelson, and Uri Simonsohn.
\newblock False-Positive Psychology: Undisclosed Flexibility in Data Collection and Analysis Allows Presenting Anything as Significant.
\newblock \emph{Psychological Science}, 22(11):1359--1366, 2011.

\end{thebibliography}
\end{document}